\documentclass[runningheads]{llncs}
\usepackage[utf8]{inputenc}
\usepackage{graphicx}
\usepackage{amsmath,amssymb} 
\usepackage{color}
\usepackage[width=122mm,left=12mm,paperwidth=146mm,height=193mm,top=12mm,paperheight=217mm]{geometry}
\usepackage{comment}
\usepackage{color}
\usepackage[font=small,skip=5pt]{caption}

\usepackage{tikz}

\newcommand{\specialcell}[2][c]{%
  \begin{tabular}[#1]{@{}c@{}}#2\end{tabular}}

\graphicspath{{./figures/}}

\begin{document}
\pagestyle{headings}
\mainmatter

\title{Now you see me: evaluating performance in long-term visual tracking} 

\author{Alan Lukežič$^1$, Luka Čehovin Zajc$^1$, Tomáš Vojíř$^2$, \\ Jiří Matas$^2$ and Matej Kristan$^1$
}

\institute{$^1$Faculty of Computer and Information Science, University of Ljubljana, Slovenia \\
$^2$Faculty of Electrical Engineering, Czech Technical University in Prague, Czech Republic}

\authorrunning{Lukežič A., Čehovin Zajc L., Vojíř T., Matas J., Kristan M.}

\maketitle

\begin{abstract}
We propose a new long-term tracking performance evaluation methodology and present a new challenging dataset of carefully selected sequences with many target disappearances. 
We perform an extensive evaluation of six long-term and nine short-term state-of-the-art trackers, using new performance measures, suitable for evaluating long-term tracking -- tracking precision, recall and F-score. The evaluation shows that a good model update strategy and the capability of image-wide re-detection are critical for long-term tracking performance. We integrated the methodology in the VOT toolkit to automate experimental analysis and benchmarking and to facilitate the development of long-term trackers.
\end{abstract}

\section{Introduction}  \label{sec:introduction}

The field of visual object tracking has significantly advanced over the last decade. The progress has been fostered by the emergence of standardized datasets and performance evaluation protocols~\cite{otb_cvpr2013,alov_pami2014,templecolor_tip2015,kristan_vot_tpami2016,MOTChallenge2015} in combination with tracking challenges \cite{kristan_vot2017,MOTChallenge2015}. Dominant single-target tracking benchmarks~\cite{otb_pami2015,alov_pami2014,templecolor_tip2015,kristan_vot_tpami2016} focus on short-term trackers. Over the time, this which lead to the development of short-term trackers that cope well with significant appearance and motion changes and are robust to short-term occlusions. Several recent publications~\cite{uav_benchmark_eccv2016,moudgil2017long,tao2017tracking} have shown that short-term trackers fare poorly on very long sequences, since the localization errors and updates gradually deteriorate their visual model, leading to drift and failure. Failure recovery, however, is primarily addressed in long-term trackers.

\begin{figure}[t]
\begin{center}
	\includegraphics[width=\linewidth]{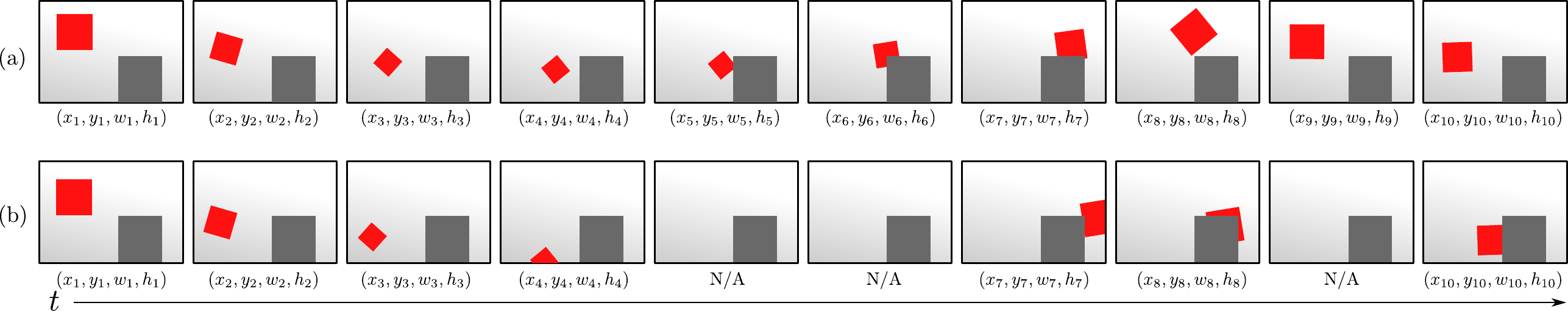}
\end{center}
   \caption{Differences between short-term and long-term tracking. (a) In short-term tracking, the target, a red box, may move and change appearance, but  it is always at least partially visible. (b) In long-term tracking, the box may disappear from the view or be fully occluded by other objects for long periods of time. Within these periods, the state of the object is not defined and should not be reported by the tracker.} 
\label{fig:long-term-tracking} 
\end{figure}

Long-term tracking does not just refer to the sequence length, as stated in~\cite{uav_benchmark_eccv2016,moudgil2017long,tao2017tracking}, but also to the sequence properties (number of target disappearances, etc.) and the type of tracking output expected. As shown in Figure~\ref{fig:long-term-tracking}, in a short-term tracking setup the object is always in the camera field of view, but not necessarily fully visible. The tracker thus reports the position of a target, which is present in each frame. In a long-term setup, the object may leave the field of view or become fully occluded for a long period. A long-term tracker is thus required to report the target position in each frame and provide a confidence score of target presence. A crucial difference to short-term tracking is thus the re-detection capability, which requires fundamentally different visual model adaptation mechanisms. These long-term aspects have been explored far less than the short-term counterparts due to lack of benchmarks and performance measures probing long-term capabilities. This is the focus of our work.

The paper makes the following contributions.
(1) A new long-term tracking performance evaluation methodology which introduces new performance measures to evaluate trackers: {\em tracking} precision, recall and F-score.
(2) We constructed a new dataset of carefully selected sequences with many target disappearances that emphasize long-term tracking properties. Sequences are annotated with ten visual attributes which enable in-depth analysis of trackers.
(3) We provide a new short-term/long-term taxonomy.  We experimentally show that re-detection capability is critically important for long-term tracking performance. 
(4) We performed an extensive evaluation of many long-term and short-term trackers in the long-term tracking scenario together with an analysis of their speed. All trackers, performance measures and evaluation protocol have been integrated into the VOT toolkit~\cite{kristan_vot_tpami2016}, to automate experimental analysis and benchmarking and facilitate development of long-term trackers. The dataset, all the trackers as well as the changes to the toolkit will be made publicly available.

\section{Related work}  \label{sec:related_work}

Performance evaluation in single-object tracking has primarily focused on short-term trackers~\cite{otb_pami2015,kristan_vot_tpami2016,templecolor_tip2015,alov_pami2014}. The currently widely-used methodologies originate from three benchmarks, OTB~\cite{otb_cvpr2013,otb_pami2015}, VOT~\cite{kristan_vot2013,kristan_vot_tpami2016} and ALOV~\cite{alov_pami2014} which primarily differ in the dataset construction, performance measures and evaluation protocols.

Benchmarks like~\cite{otb_pami2015,alov_pami2014,templecolor_tip2015} propose large datasets, reasoning that quantity reduces the variance in performance estimation. Alternatively, the longest-running benchmark~\cite{kristan_vot_tpami2016} argues that quantity does not necessarily mean quality and promotes moderate-sized datasets with carefully chosen diverse sequences for fast an informative evaluation. Several works have focused on specific tracking setups. Mueller et. al.~\cite{uav_benchmark_eccv2016} proposed the UAV123 dataset for tracking from aerial vehicles. Galoogahi et al.~\cite{Galoogahi_2017_ICCV} introduced a high-frame-rate dataset to analyze trade-offs between tracker speed and robustness. \v{C}ehovin et al.~\cite{cehovin_iccv2017} proposed a dataset with an active camera view control using omni directional videos for accurate tracking analysis as a function camera motion attributes. The target never leaves the field of view in these datasets, making them unstable for long-term tracking properties evaluation. 

Many performance measures have been explored in tracking~\cite{cehovin_tip2016}. All dominant short-term performance measures~\cite{otb_pami2015,alov_pami2014,kristan_vot_tpami2016} are based on the overlap (intersection over union) between the ground truth bounding boxes and tracker predictions, but significantly differ in the use. ALOV~\cite{alov_pami2014} uses the F-measure computed at overlap threshold of 0.5. OTB~\cite{otb_pami2015} avoids the threshold by computing the average overlap over the sequences as the primary measure. The VOT~\cite{kristan_vot_tpami2016} 
resets the tracker once the overlap drops to zero, and proposes to measure robustness by the number of times the tracker was reset, the accuracy by average overlap during successful tracking periods and an expected average overlap on a typical short-term sequence. These measures do not account for tracker ability to report target absence and are therefore not suitable for long-term tracking.

A few papers have recently addressed the datasets focusing on long-term performance evaluation. Tao et al.~\cite{tao2017tracking} created artificial long sequences by repeatedly playing shorter sequences forward and backward. Such a dataset exposes the problem of gradual drift in short-term trackers, but does not fully expose the long-term abilities since the target never leaves the field of view. Mueller et al.~\cite{uav_benchmark_eccv2016} proposed UAV20L dataset of twenty long sequences with target frequently exiting and re-entering the scene, but used it to evaluate mostly short-term trackers. 
A dataset with many cases of fully occluded and absent target has been recently proposed in~\cite{moudgil2017long}. Unfortunately, the large number of target disappearances was obtained by significantly increasing the sequence length, which significantly increases the storage requirements. To cope with this, a very high video compression is applied, thus sacrificing the image quality. 
   
In the absence of clear a long-term tracking definition, much less attention has been paid to long-term performance measures. The UAV20L~\cite{uav_benchmark_eccv2016} and~\cite{moudgil2017long} apply the short-term average overlap measure~\cite{otb_pami2015}, which does not account for situation when the tracker reports target absence and favors the trackers that report target positions at every frame. Tao et al.~\cite{tao2017tracking} adapted this measure by specifying an overlap $1$ when the tracker correctly predicts the target absence. Nevertheless, this value is not "calibrated" with the tracker accuracy when the target is visible, which skews the overlap-based measure. Furthermore, reducing the actual tracking accuracy and failure detection to a single overlap score significantly limits its the insight it brings.

\section{Short-term/Long-term tracking spectrum}\label{sec:long-term}

A long-term tracker is required to handle target disappearance and reappearance (Figure~\ref{fig:long-term-tracking}). Relatively few published trackers fully address the long-term requirements, and yet some short-term trackers address them partially. We argue that trackers should not be simply classified as short-term or long-term, but they rather cover an entire short-term--long-term \textit{spectrum}. The following taxonomy is used in our experimental section for accurate performance analysis.

\begin{enumerate}
\item  {\bf Short-term tracker} ($\mathrm{ST}_0$). The target position is reported at each frame. The tracker does not implement target re-detection and does not explicitly detect occlusion. Such trackers are likely to fail on first occlusion as their representation is affected by any occluder. 
\item {\bf Short-term tracker with conservative updating} ($\mathrm{ST}_1$). The target position is reported at each frame. Target re-detection is not implemented, but tracking robustness is increased by selectively updating the visual model depending on a tracking confidence estimation mechanism.
\item {\bf Pseudo long-term tracker} ($\mathrm{LT}_0$). The target position is not reported in frames when the target is not visible. The tracker does not implement explicit target re-detection but uses an internal mechanism to identify and report tracking failure.
\item {\bf Re-detecting long-term tracker} ($\mathrm{LT}_1$). The target position is not reported in frames when the target is not visible. The tracker detects tracking failure and implements explicit target re-detection. 
\end{enumerate}

The $\mathrm{ST}_0$ and $\mathrm{ST}_1$ trackers are what is commonly considered a short-term tracker. Typical representatives from $\mathrm{ST}_0$ are KCF~\cite{henriques2015tracking}, SRDCF~\cite{srdcf_iccv2015} and CSRDCF~\cite{Lukezic_CVPR_2017}. MDNet~\cite{mdnet_cvpr2016} and ECO~\cite{danelljan_iccv2015_convolutional} are current state-of-the art trackers from $\mathrm{ST}_1$. Many short-term trackers can be trivially converted into pseudo long-term trackers ($\mathrm{LT}_0$) by using their visual model similarity scores at the reported target position. While straightforward, this offers means to evaluate short-term trackers in the long-term context.

The level $\mathrm{LT}_1$ trackers are the most sophisticated long-term trackers, in that they cover all long-term requirements.
These trackers typically combine two components, a short-term tracker and a detector, and implement an algorithm for their interaction. The $\mathrm{LT}_1$ trackers originate from two main paradigms introduced by TLD~\cite{kalal_pami} and Alien~\cite{Pernici2013}, with modern examples CMT~\cite{CMT_CVPR2015}, Matrioska~\cite{Maresca2013}, MUSTER~\cite{muster_cvpr2015}, LCT~\cite{LCT_CVPR2015}, PTAV~\cite{ptav_iccv2017}, and FCLT~\cite{fclt_arxiv}. Interestingly, two recently published trackers LCT~\cite{LCT_CVPR2015} and  PTAV~\cite{ptav_iccv2017}, that
perform well in short-term evaluation benchmarks (OTB50~\cite{otb_cvpr2013} and OTB100~\cite{otb_pami2015}),  are presented as long-term trackers ~\cite{ptav_iccv2017,LCT_CVPR2015}, but experiments in Section~\ref{sec:trackers} show they are in the LT$_0$ class.

\section{Long-term tracking performance measures}  \label{sec:methodology}

A long-term tracking performance measure should reflect the localization accuracy, but unlike short-term measures, it should also capture the accuracy of target absence prediction as well as target re-detection capabilities. These properties are quantified by the \textit{precision} and \textit{recall} measures widely used in detection literature~\cite{everingham2010pascal}, and they are the basis for the proposed long-term performance measures.

Let $G_t$ be the ground truth target pose, let $A_t(\tau_\theta)$ be the pose predicted by the tracker, $\theta_t$ the prediction certainty score at time-step $t$ and $\tau_\theta$ be a classification threshold. If the target is absent, the ground truth is an empty set, i.e., $G_t=\emptyset$. 
Similarly, if the tracker did not predict the target or 
 the prediction certainty score is below a classification threshold i.e., $\theta_t < \tau_\theta$,
the output is $A_t(\tau_\theta)=\emptyset$. 
The agreement between the ground truth and prediction is specified by their intersection over union $\Omega(A_t(\tau_\theta), G_t)$\footnote{The output of $\Omega(\cdot, \cdot)$ is 0 if any of the two regions is $\emptyset$.}. In detection literature, the prediction matches the ground truth if the overlap $\Omega(A_t(\tau_\theta), G_t)$ exceeds a  threshold $\tau_\Omega$. Given the two thresholds $(\tau_\theta, \tau_\Omega)$, the precision $Pr$ and recall $Re$ are defined as
\begin{eqnarray} \label{eq:pr_re_general}
Pr(\tau_\theta, \tau_\Omega) = | \{ t : \Omega(A_t(\tau_\theta), G_t) \geq \tau_\Omega \} | / N_p, \\
Re(\tau_\theta, \tau_\Omega) = | \{ t : \Omega(A_t(\tau_\theta), G_t) \geq \tau_\Omega \} | / N_g,
\end{eqnarray}
where $|\cdot|$ is the cardinality, $N_g$ is the number of frames with $G_t\neq\emptyset$ and $N_p$ is the number of frames with existing prediction, i.e. $A_t(\tau_\theta) \neq \emptyset$.

In detection literature, the overlap threshold is set to $0.5$ or higher, while recent work~\cite{kristan_vot_tpami2016} has demonstrated that such threshold is over-restrictive and does not clearly indicate a tracking failure in practice. A popular short-term performance measure~\cite{otb_cvpr2013}, for example, addresses this by averaging performance over various thresholds, which was shown in~\cite{cehovin_tip2016} to be equal to the average overlap. Using the same approach, we reduce the precision and recall to a single threshold by integrating over $\tau_\Omega$, i.e., 

\begin{eqnarray} \label{eq:pr_re}
Pr(\tau_\theta) = \int_0^1 Pr(\tau_\theta, \tau_\Omega) d\tau_\Omega = \frac{1}{N_p} \sum_{t \in \{ t : A_t(\theta_t) \neq \emptyset \}  } \Omega(A_t(\theta_t), G_t), \\
Re(\tau_\theta) = \int_0^1 Re(\tau_\theta, \tau_\Omega) d\tau_\Omega = \frac{1}{N_g} \sum_{t \in \{ t : G_t \neq \emptyset \}  } \Omega(A_t(\theta_t), G_t).
\end{eqnarray}

We call $Pr(\tau_\theta)$ {\em tracking} precision and $Re(\tau_\theta)$ {\em tracking} recall to distinguish them from their detection counterparts. Detection-like precision/recall plots can be drawn to analyze the tracking as well as detection capabilities of a long-term tracker (Figure~\ref{fig:average_f_pr_re}). Similarly, a standard trade-off between the precision and recall can be computed in form of a {\em tracking} F-score~\cite{everingham2010pascal}
\begin{eqnarray} \label{eq:f_measure}
F(\tau_\theta) = 2 Pr(\tau_\theta) Re(\tau_\theta) / (Pr(\tau_\theta) + Re(\tau_\theta)),
\end{eqnarray}
\noindent and visualized by the F-score plots (Figure~\ref{fig:average_f_pr_re}). Our primary score for ranking long-term trackers is therefore defined as the highest F-score on the F-score plot, i.e., taken at the tracker-specific optimal threshold. This avoids manually-set thresholds in the primary performance measure.
 
Note that the proposed primary measure (\ref{eq:f_measure}) for the long-term trackers is consistent with the established short-term tracking methodology. Consider an $\mathrm{ST}_0$ short-term tracking scenario: the target is always (at least partially) visible and the target position is predicted at each frame with equal certainty. In this case our F-measure (\ref{eq:f_measure}) reduces to the average overlap, which is a standard measure in short-term tracking~\cite{otb_cvpr2013,kristan_vot_tpami2016}.

\section{The long-term dataset (LTB35)}  \label{sec:dataset}

Table~\ref{tab:datasets} quantifies the long-term statistics of the common short-term and existing long-term tracking datasets. Target disappearance is missing in the standard short-term datasets except for UAV123 which contains on average less than one full occlusion per sequence. This number increases four-fold in UAV20L~\cite{uav_benchmark_eccv2016} long-term dataset. The recent TLP~\cite{moudgil2017long} dataset increases the number of target disappearances by an order of magnitude, but at a cost of increasing the dataset size in terms of the number of frames by more than an order of magnitude, i.e.  target disappearance events are less frequent in TLP~\cite{moudgil2017long} than in UAV20L~\cite{uav_benchmark_eccv2016}, see Table~\ref{tab:datasets}. Moreover, the videos are heavily compressed with many artifacts that affect tracking.

\begin{table}[h]
\begin{center}
\caption{Datasets -- comparison of long-term properties: the number of sequences, the total number of frames, the number of target disappearances (DSP), the average length of disappearance interval (ADL), the average number of disappearances in sequence (ADN). The first four datasets are short-term with virtually no target disappearances, the last column shows the properties of the proposed dataset.}
\label{tab:datasets}
\scalebox{.80}{
\begin{tabular}{l | c | c | c | c | c | c | c}
\hline
{\bf Dataset} & ALOV300~\cite{alov_pami2014} & OTB100~\cite{otb_pami2015} & VOT2017~\cite{kristan_vot2017} & UAV123~\cite{uav_benchmark_eccv2016} & UAV20L~\cite{uav_benchmark_eccv2016} & TLP~\cite{moudgil2017long} & LTB35 (ours) \\
\hline
{\bf \# sequences} & 315 & 100 & 60 & 123 & 20 & 50 & 35 \\
{\bf Frames} & 89364 & 58897 & 21356 & 112578 & 58670 & 676431 & 146847 \\
{\bf DSP} & 0 & 0 & 0 & 63 & 40 & 316 & 433 \\
{\bf ADL} & 0 & 0 & 0 & 42.6 & 60.2 & 64.1 & 40.6 \\
{\bf ADN } & 0 & 0 & 0 & 0.5 & 2 & 6.3 & 12.4 \\
\hline
\end{tabular}
}
\end{center}
\end{table}

In the light of the limitations of the existing datasets, we created a new long-term dataset. We  followed the VOT~\cite{kristan_vot_tpami2016} dataset construction paradigm which states that the datasets should be kept moderately large and manageable, but rich in attributes relevant to the tested tracker class. We started by including all sequences from UAV20L since they contain a moderate occurrence  of occlusions and potentially difficult to track small targets. Three sequences were taken from~\cite{kalal_pami}. We collected six additional sequences from Youtube. The sequences contain larger targets with numerous disappearances. To further increase the number of target disappearances per sequence, we have utilized the recently proposed omni-directional AMP dataset~\cite{cehovin_iccv2017}. Six additional challenging sequences were generated from this dataset by controlling the camera such that the target repeatedly entered the field of view from one side and left it at the other. 

\begin{figure}[h]
\begin{center}
	\includegraphics[width=\linewidth]{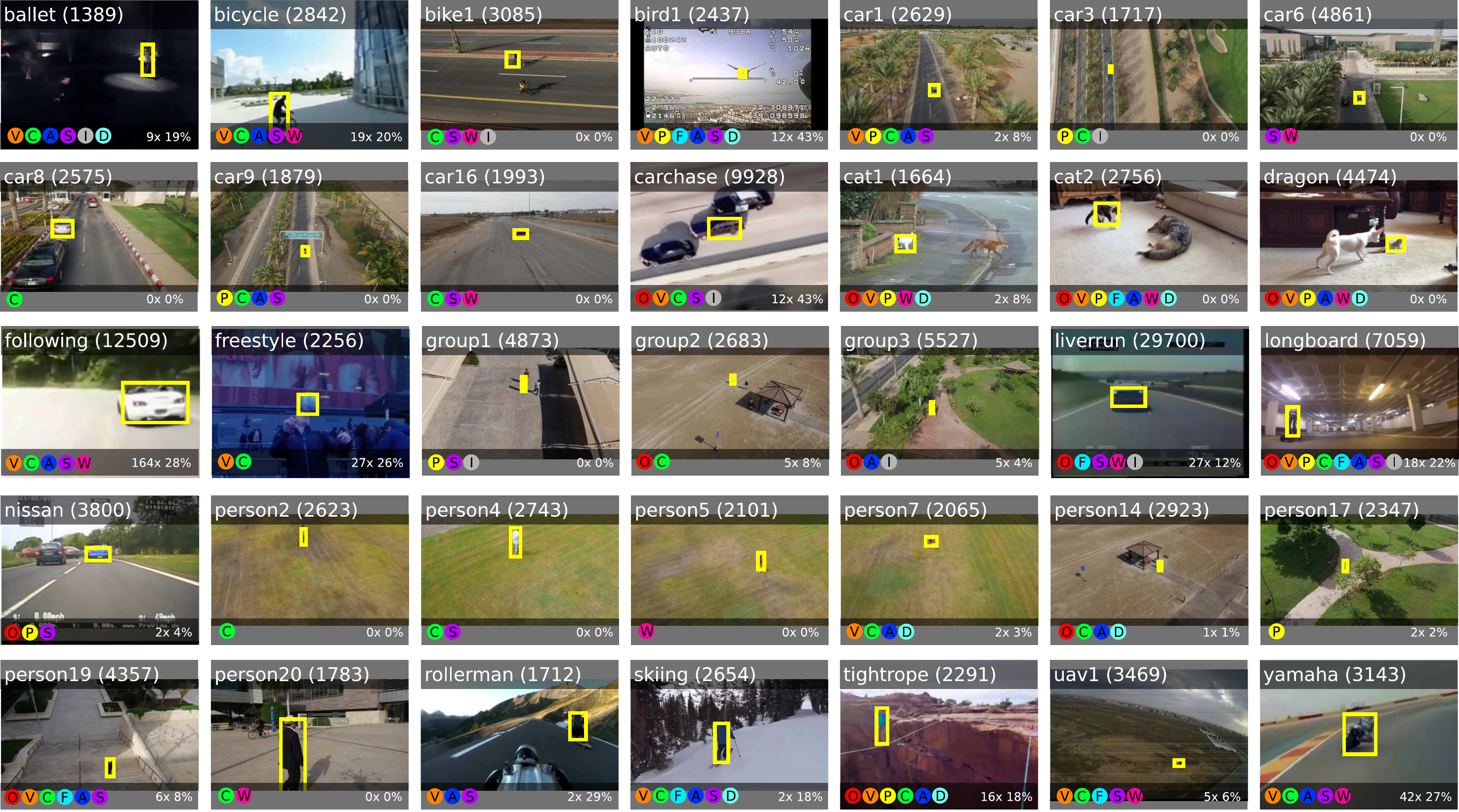}
\end{center}
   \caption{The LTB35 dataset -- a  frame selected from each sequence. Name and length  (top),  number of disappearances and percentage of frames without target (bottom right). Visual attributes (bottom left): (O) Full occlusion, (V) Out-of-view, (P) Partial occlusion, (C) Camera motion, (F) Fast motion, (S) Scale change, (A) Aspect ratio change, (W) Viewpoint change, (I) Similar objects, (D) Deformable object.} 
\label{fig:sequences} 
\end{figure}

The targets were annotated by axis-aligned bounding-boxes. Each sequence is annotated by ten visual attributes: full occlusion, out-of-view motion, partial occlusion, camera motion, fast motion, scale change, aspect ratio change, viewpoint change, similar objects, and deformable object.
The LTB35 thus contains 35 challenging sequences of various objects (persons, car, motorcycles, bicycles, animals, etc.) with the total length of $146847$ frames. Sequence resolutions range between $1280 \times 720$ and $290 \times 217$. Each sequence contains on average 12 long-term target disappearances, each lasting on average 40 frames. An overview of the dataset is shown in Figure~\ref{fig:sequences}.

\section{Experimental evaluation}  \label{sec:experiments}
 
\subsection{Evaluation protocol}

A tracker is evaluated on a dataset of several sequences by initializing on the first frame of a sequence and run until the end of the sequence without re-sets. The precision-recall graph (\ref{eq:pr_re}) is calculated on each sequence and averaged into a single plot. This guarantees that the result is not dominated by extremely long sequences. The F-measure plot (\ref{eq:f_measure}) is computed from the average precision-recall plot and used to compute the primary ranking score. The evaluation protocol along with plot generation was implemented in the VOT~\cite{kristan_vot_tpami2016} toolkit to automate experiments and thus reduce potential human errors.

\subsection{Evaluated trackers}  \label{sec:trackers}

\begin{table}[t!]
\begin{center}
\caption{Evaluated trackers. All trackers are characterized by the short-term component and their confidence score, long-term trackers are also characterized by the detector type   and its interaction with the short-term component. We also summarize model update and search strategies. Trackers marked by $^*$ were published as $LT_1$, but did not pass the re-detection test.}

\label{tab:trackers}
\scalebox{.8}{
\begin{tabular}{l | c | c | c | c | c | c}
\hline
{\bf Tracker} & {\bf S-L} & {\bf Detector} & \specialcell{{\bf Short-term} \\ {\bf component}} & \specialcell{{\bf Interaction} \\ {\bf Score}} & {\bf Update} & {\bf Search} \\
\hline
TLD~\cite{kalal_pami} & $LT_1$ & \specialcell{Random \\fern} & Flow & \specialcell{P-N learning \\ Score: conser. sim.} & \specialcell{Positive, \\negative samp.} & \specialcell{Entire image \\ (cascade)} \\
\hline
MUSTER~\cite{muster_cvpr2015} & $LT_1$ & \specialcell{Keypoints \\ (SIFT)} & CF & \specialcell{F-B, RANSAC \\ Score: max. corr.} & \specialcell{ST: every frame \\ LT: when confident} & \specialcell{Entire image \\ (keypoint matching)} \\
\hline
FCLT~\cite{fclt_arxiv} & $LT_1$ & CF (reg.) & CF (reg.) & \specialcell{Resp. thresh., \\Score: resp. quality} & \specialcell{ST: when confident \\ LT: mix ST + LT} & \specialcell{Entire image \\ (correlation + motion)} \\
\hline
CMT~\cite{CMT_CVPR2015} & $LT_1$ & \specialcell{Keypoints \\(static)} & \specialcell{Keypoints \\(flow)} & \specialcell{F-B, clustering, \\correspondencies\\ Score: \# keypoints} & \specialcell{ST: always \\ LT: never} & \specialcell{Entire image \\ (keypoint matching)} \\
\hline
PTAV$^*$~\cite{ptav_iccv2017} & $LT_0$ & \specialcell{Siamese \\network} & \specialcell{CF \\ (fDSST)} & \specialcell{Conf. thresh,\\ const. verif. interval \\ Score: CNN score} & \specialcell{ST: always, \\ LT: never} & \specialcell{Search window \\ (enlarged region)} \\
\hline
LCT$^*$~\cite{LCT_CVPR2015} & $LT_0$ & \specialcell{Random \\fern} & CF & \specialcell{k-NN, resp. thresh.\\ Score: max. corr.} & \specialcell{When \\ confident} & \specialcell{Search window \\ (enlarged region)} \\
\hline
SRDCF~\cite{srdcf_iccv2015} & $ST_0$ & - & CF & \specialcell{- \\ Score: max. corr.} & \specialcell{Always \\(exp. forget.)} & \specialcell{Search window \\ (enlarged region)} \\
\hline
ECO~\cite{DanelljanCVPR2017} & $ST_1$ & - & \specialcell{CF \\(deep f.)} & \specialcell{- \\ Score: max. corr.} & \specialcell{Always \\(clustering)} & \specialcell{Search window \\ (enlarged region)} \\
\hline
ECOhc~\cite{DanelljanCVPR2017} & $ST_1$ & - & CF & \specialcell{- \\ Score: max. corr.} & \specialcell{Always \\(clustering)} & \specialcell{Search window \\ (enlarged region)} \\
\hline
KCF~\cite{henriques2015tracking} & $ST_0$ & - & CF & \specialcell{- \\ Score: max. corr.} & \specialcell{Always \\(exp. forget.)} & \specialcell{Search window \\ (enlarged region)} \\
\hline
CSRDCF~\cite{Lukezic_CVPR_2017} & $ST_0$ & - & CF & \specialcell{- \\ Score: max. corr.} & \specialcell{Always \\(exp. forget.)} & \specialcell{Search window \\ (enlarged region)} \\
\hline
BACF~\cite{BACF_ICCV2017} & $ST_0$ & - & CF & \specialcell{- \\ Score: max. corr.} & \specialcell{Always \\(exp. forget.)} & \specialcell{Search window \\ (enlarged region)} \\
\hline
SiamFC~\cite{siamfc_eccv16} & $ST_1$ & - & CNN & \specialcell{- \\ Score: max. corr.} & Never & \specialcell{Search window \\ (enlarged region)} \\
\hline
MDNet~\cite{mdnet_cvpr2016} & $ST_1$ & - & CNN & \specialcell{- \\ Score: CNN score} & \specialcell{When confident \\ (hard negatives)} & \specialcell{Random \\ sampling} \\
\hline
CREST~\cite{crest_ICCV17} & $ST_0$ & - & CNN & \specialcell{- \\ Score: max. corr.} & \specialcell{Always \\ (backprop)} & \specialcell{Search window \\ (enlarged region)} \\
\hline
\end{tabular}
}
\end{center}
\end{table}

An extensive collection of top-performing trackers was complied to cover the short-term--long-term spectrum. In total fifteen trackers, summarized in Table~\ref{tab:trackers} and Figure~\ref{fig:circle_plot}, were evaluated. 
We included six long-term state-of-the-art trackers with publicly available source code: (i) TLD~\cite{kalal_pami}, which uses optical flow for short-term component and normalized-cross-correlation for detector and a P-N learning framework for detector update. (ii) LCT~\cite{LCT_CVPR2015} and (iii) MUSTER~\cite{muster_cvpr2015} that
use a discriminative correlation filter for the short-term component and random ferns and keypoints, respectively, for the detector. (iv) PTAV~\cite{ptav_iccv2017} that uses a correlation filter for short-term component and a CNN retrieval system~\cite{sint_cvpr16} for detector. (v) FCLT~\cite{fclt_arxiv} that uses a correlation filter for both, short-term component and detector. (vi) CMT~\cite{CMT_CVPR2015} that uses optical flow for short-term component and key-points for detector. These trackers further vary in the frequency and approach for model updates (see Table~\ref{tab:trackers}).

In addition to the selected long-term trackers, we have included recent state-of-the art short-term trackers. A standard discriminative correlation filter KCF~\cite{henriques2015tracking}, four recent advanced versions SRDCF~\cite{srdcf_iccv2015}, CSRDCF~\cite{Lukezic_CVPR_2017}, BACF~\cite{BACF_ICCV2017}, ECOhc~\cite{DanelljanCVPR2017} and the top-performer on the OTB~\cite{otb_cvpr2013} benchmark ECO~\cite{DanelljanCVPR2017}. Two state-of-the-art CNN-based top-performers from the VOT~\cite{kristan_vot2016} benchmark SiamFC~\cite{siamfc_eccv16} and MDNet~\cite{mdnet_cvpr2016}  and a top-performing CNN-based tracker CREST~\cite{crest_ICCV17} were included as well. All these short-term trackers were modified to be LT$_0$ compliant. A reasonable score was identified in each tracker and used as the uncertainty score to detect tracking failure. All trackers were integrated in the VOT~\cite{kristan_vot_tpami2016} toolkit for automatic evaluation.

\subsubsection{\bf Re-detection experiment.}  \label{sec:redetection-experiment}
An experiment was designed to position the tested trackers on the LT/ST spectrum, and in particular to verify their re-detection capability. Artificial sequences were generated from the initial frame of each sequence in our dataset, in these sequences the target appearance was kept constant to emphasize the re-detection mechanism performance.

\begin{figure}[!htb]
\begin{minipage}[t]{0.49\textwidth}
\begin{center}
	\includegraphics[width=\linewidth]{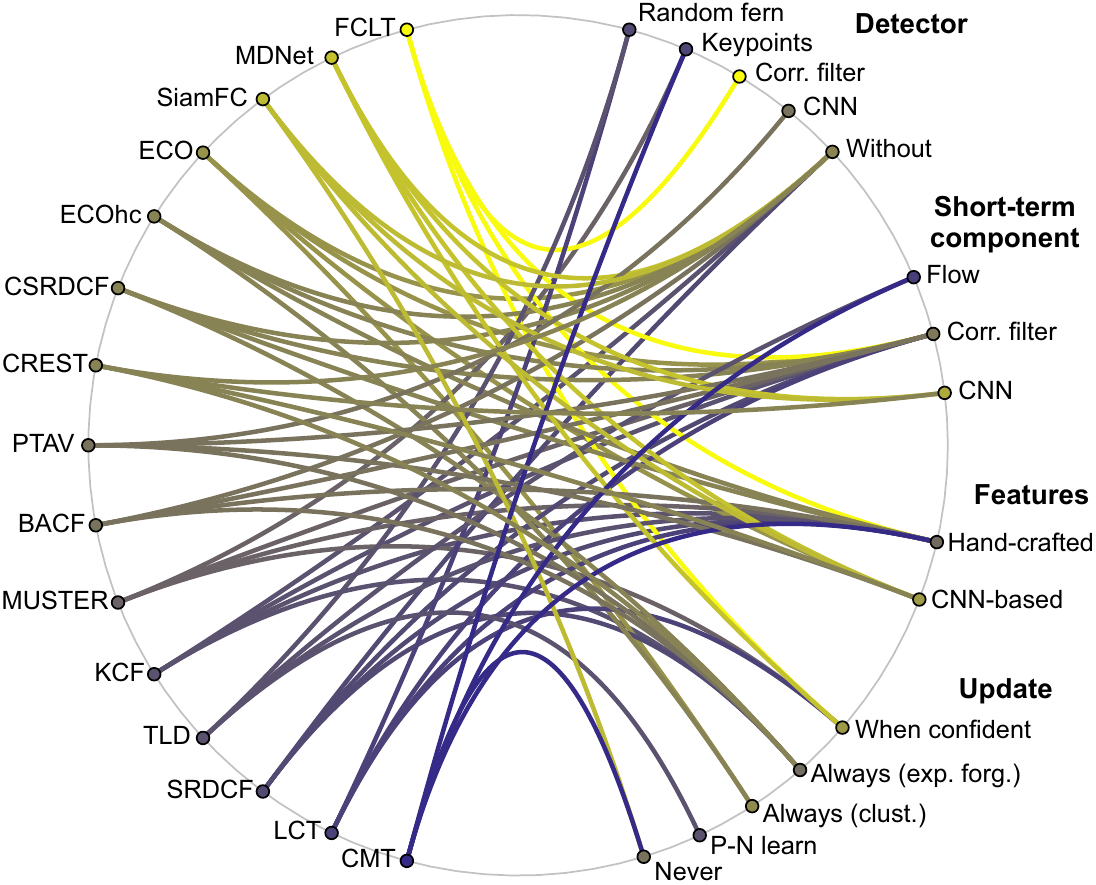}
\end{center}
   \caption{Structure of the trackers. The links characterize tracker components. Color codes performance on the LTB35 benchmark, yellow - best, blue worst. 
} 
\label{fig:circle_plot} 
\end{minipage}\hfill
\begin{minipage}[t]{0.49\textwidth}
\begin{center}
	\includegraphics[width=\linewidth]{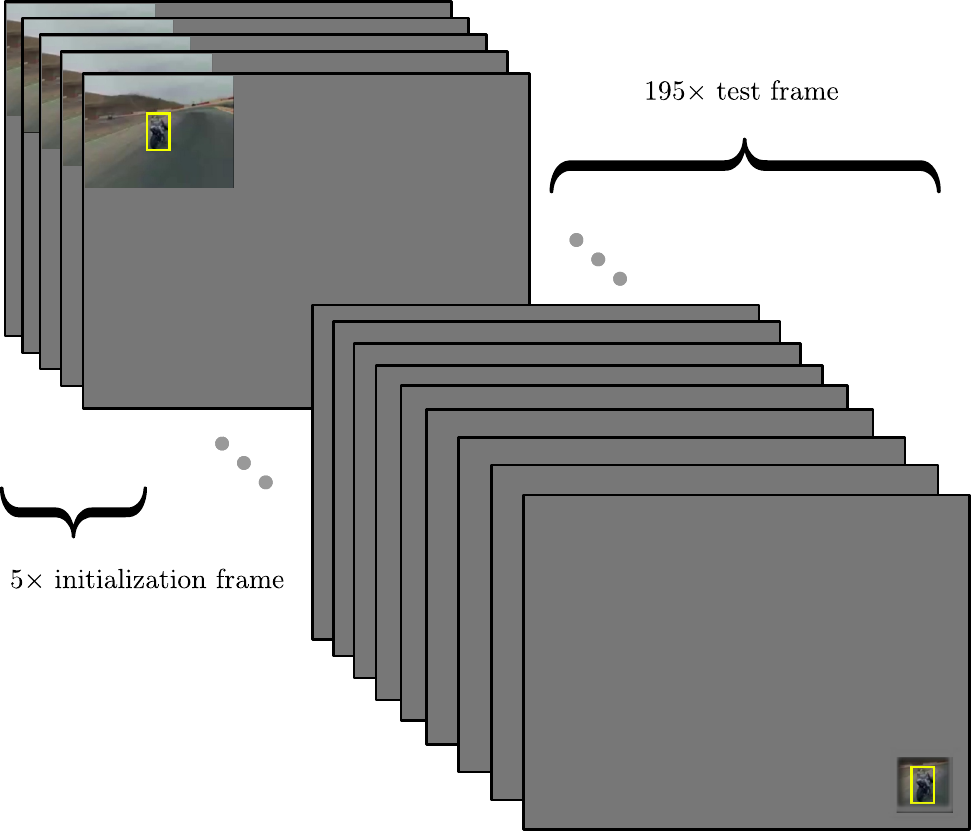}
\end{center}
   \caption{Re-detection experiment -- the artificially created sequence structure by repetition, padding and target displacement. For more, see 
   text.} 
\label{fig:redetection-image} 
\end{minipage}
\end{figure}

An initial frame of a sequence was padded with zeros right and down to the three times original size (Figure~\ref{fig:redetection-image}). This frame was repeated for the first five frames in the artificial sequence. For the remainder of the frames, the target was cropped from the initial image and placed in the bottom right corner of the frame. A tracker was initialized in the first frame and we measured the number of frames required to re-detect the target after position change. 

Results are summarized in Table~\ref{tab:redetection}. 
The trackers MDNet, BACF, ECO, ECOhc, SRDCF, SiamFC, CREST, CSRDCF and KCF never re-detected the target, which confirms their short-term design. The only tracker that successfully re-detected the target was FCLT, while MUSTER, CMT and TLD were successful in most sequences -- this result classifies them as $LT_1$ trackers. The difference in detection success come from the different detector design. FCLT and TLD both train template-based detectors. The improvement of the FCLT likely comes from the efficient discriminative filter training framework of the FCLT detector. The keypoint-based detectors in MUSTER and CMT are similarly efficient, but require sufficiently well textured targets. Interestingly the re-detection is imminent for Muster, CMT and TLD, while FCLT requires on average 79 frames. This difference comes form the dynamic models. The Muster, CMT and TLD apply a uniform dynamic model in the detector phase over the entire image, while the FCLT applies a random walk model, that gradually increases the target search range with time.

Surprisingly two recent long-term trackers, LCT and PTAV nearly never successfully detected the target. A detailed inspection of their source code revealed that these trackers do not apply their detector to the whole image, but rather a small neighborhood of the previous target position, which makes these two trackers a pseudo long-term, i.e., $LT_0$ level. 

\begin{table}[t]
\begin{center}
\caption{Re-detection results. Success -- the number of successful re-detections in 35 sequences. Frames -- the average number of frames before successful re-detection.}
\label{tab:redetection}
\scalebox{.7}{
\begin{tabular}{l  c  c  c  c  c  c  c  c  c  c  c  c  c  c  c}
\hline
{\bf Tracker} & FCLT & MUSTER & CMT & TLD & PTAV & LCT & MDNet & BACF & ECO & ECOhc & SRDCF & SiamFC & CREST & CSRDCF & KCF  \\
\hline
{\bf Success} & 35 & 29 & 28 & 17 & 1 & 0 & 0 & 0 & 0 & 0 & 0 & 0 & 0 & 0 & 0 \\
{\bf Frames} & 79 & 0 & 1 & 0 & 35 & - & - & - & - & - & - & - & - & - & - \\
\hline
\end{tabular}
}
\end{center}
\end{table}

\subsection{Overall performance} \label{sec:overall_evaluation}

\begin{figure}[t]
\begin{center}
	\includegraphics[width=1\linewidth]{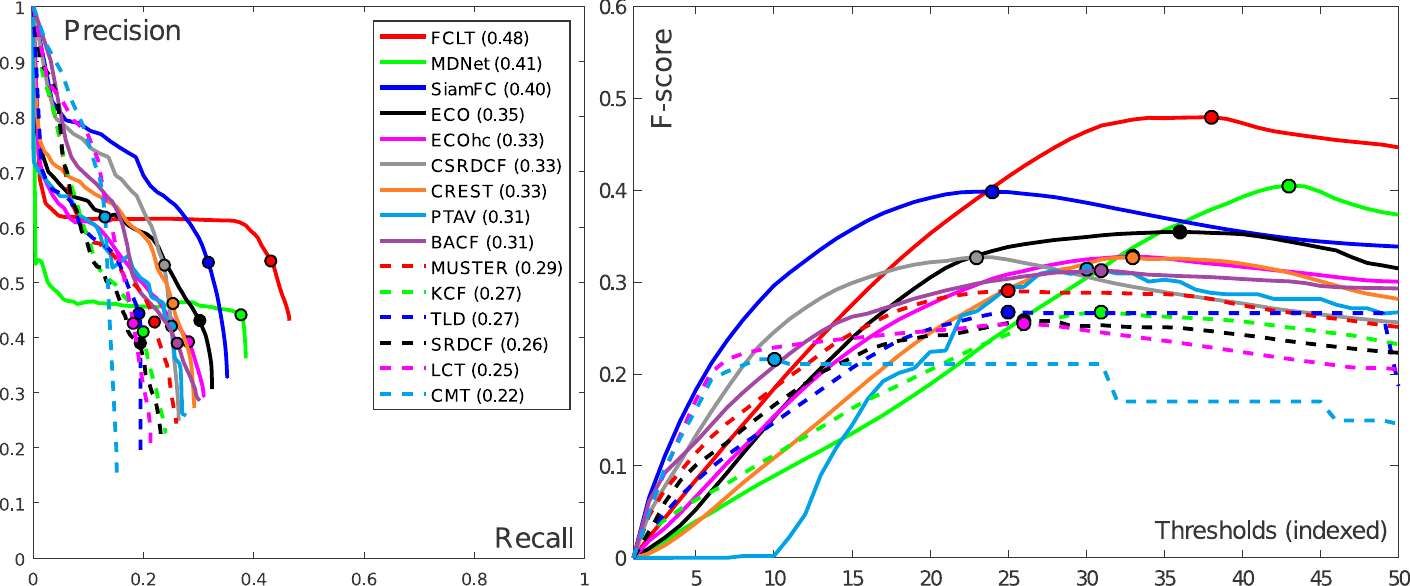}
\end{center}
   \caption{Long-term tracking performance on the LTB35 dataset. The average tracking precision-recall curves (left), the corresponding  F-score curves (right). Tracker labels are sorted according to  maximum of the F-score.} 
\label{fig:average_f_pr_re} 
\end{figure}

The overall performance on the TLB dataset is summarized in Figure~\ref{fig:average_f_pr_re}. The highest ranked is FCLT, an LT$_1$ class tracker, which uses discriminative correlation filters on hand-crafted features for short-term component as well as detector in the entire image. Surprisingly FCLT is followed by three short-term ST$_1$ class CNN-based trackers MDNet, SiamFC and ECO. These implement different mechanisms to deal with occlusion. MDNet applies very conservative updates, SiamFC does not update the model at all and ECO applies clustering-based update mechanism prevent learning from outliers. SiamFC applies a fairly large search regions, while the search region size is adapted in the MDNet by a motion model. Two long-term trackers CMT (LT$_1$) and LCT (LT$_0$) perform the worst among the tested trackers. The CMT entirely relies on keypoints, which poorly describe non-textured targets. The relatively poor performance of LCT is likely due to small search window and poor detector learning. This is supported by the fact that LCT performance is comparable to KCF, a standard correlation filter, also used as the short-term component in LCT. The performance of short-term trackers ST$_0$ class trackers does not vary significantly.

\subsection{Per-sequence evaluation}  \label{sec:sequence_evaluation}

\begin{figure}[h]
\begin{center}
	\includegraphics[width=1\linewidth]{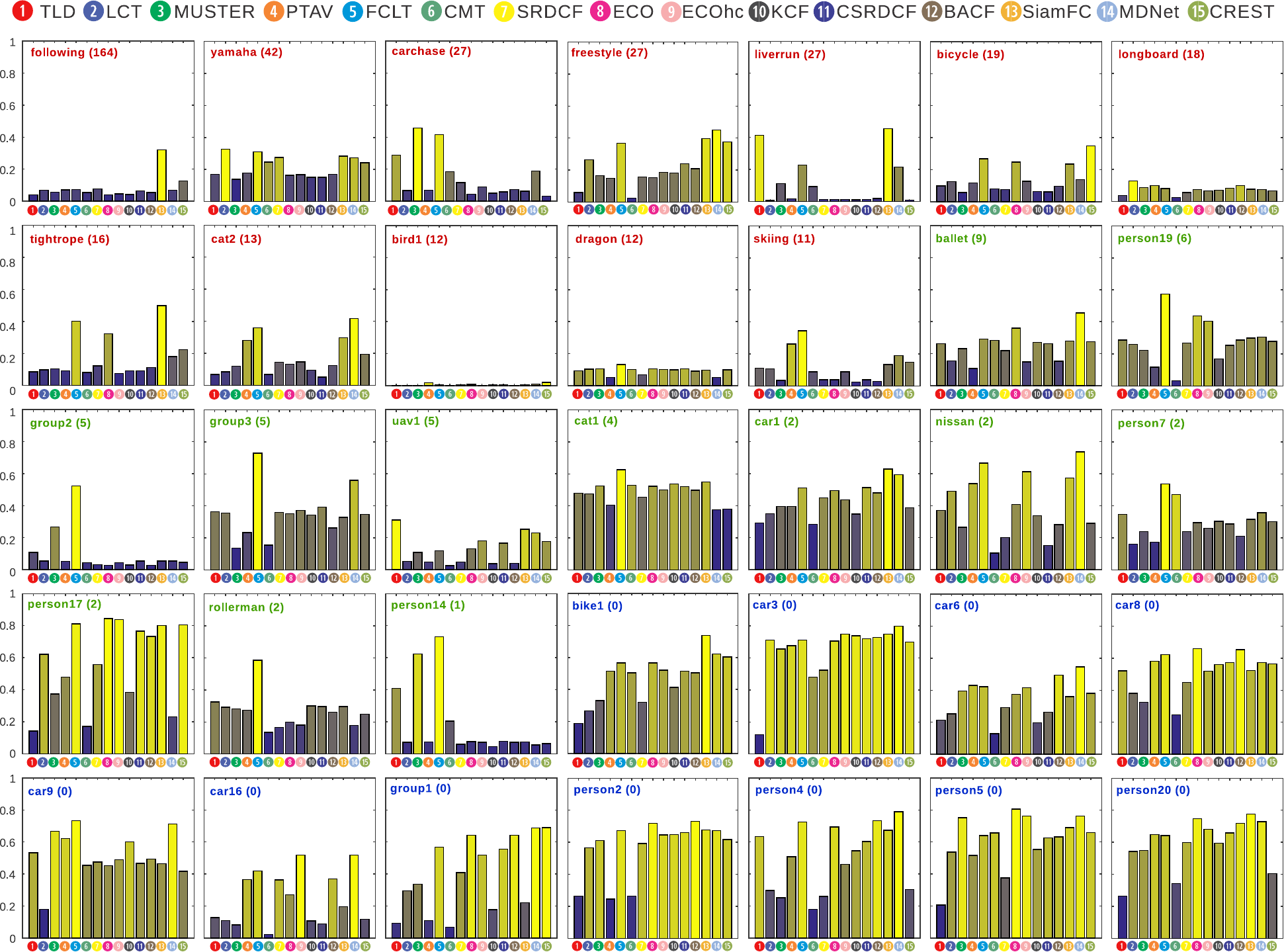}
\end{center}
   \caption{Maximum F-score of each tracker on all sequences. Sequences are sorted, left-to-right, top-to-bottom, by the  number of target disappearances, i.e. the largest number at top-left. Red label:  $>10$ disappearances, green: $1-10$, blue: no disappearance.} 
\label{fig:per_seq_average_f} 
\end{figure}

The sequences are divided into groups according to the number of target disappearances: (Group~1) over ten disappearances, (Group~2) between one and ten disappearances and (Group~3) no disappearances. Per-sequence F-scores are summarized in Figure~\ref{fig:per_seq_average_f}. 
 
\textit{Group 1 results:} Most short-term trackers performed poorly due to lack of target re-detection. Long-term trackers generally perform well, but there are differences depending on their structure. For example, the ``following'' and ``liverrun'' sequences contain cars, which only moderately change the appearance. SiamFC does not adapt the visual model and is highly successful on these sequences. The LCT generally performs poorly, except from ``yamaha'' sequence in which the target leaves and re-enters the view at the same location. Thus the poor performance of LCT is due to a fairly small re-detection range. Surprisingly some of the CNN short-term trackers perform moderately well (MDNet, CREST and SiamFC), which is likely due to highly discriminative visual features and relatively large target localization range.

\textit{Group 2 results:} Performance variation comes from a mix of target disappearance and other visual attributes. However, in ``person14'' the poor performance is related to long-lasting occlusion at the beginning, where most trackers fail. Only some of LT$_1$ class trackers (FCLT, MUSTER, and TLD) overcome the occlusion and obtain excellent performance.
 
\textit{Group 3 results:} The performance of long-term trackers does not significantly differ from short-term trackers since the target is always visible. The strength of the features and learning in visual models play a major role. These sequences are least challenging for all trackers in our benchmark.

\subsection{Attribute evaluation}  \label{sec:attribute_evaluation}

\begin{figure}[h]
\begin{center}
	\includegraphics[width=1\linewidth]{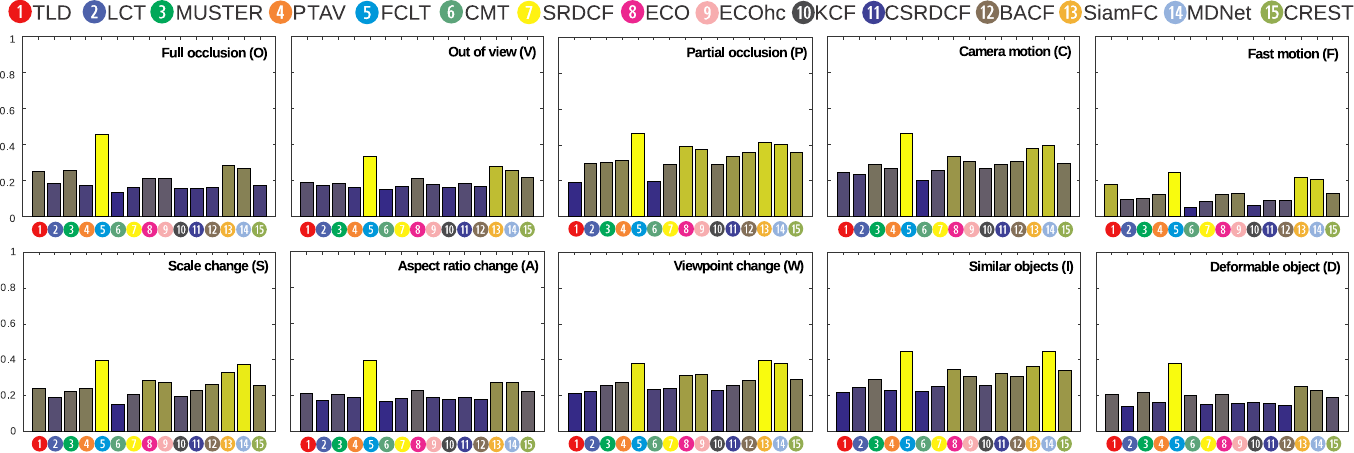}
\end{center}
   \caption{Maximum F-score averaged over overlap thresholds for the visual attributes. The most challenging attributes are fast motion, full occlusion, out-of-view and deformable object.} 
\label{fig:attributes_average_f} 
\end{figure}

Figure~\ref{fig:attributes_average_f} shows tracking performance with respect to ten visual attributes from Section~\ref{sec:dataset}. Long-term tracking is mostly characterized by performance on full occlusion and out-of-view attributes, since these require re-detection. The FCLT (LT$_1$ class) achieves top performance, which is likely due to the efficient learning of the detector component. The other LT$_1$ trackers, MUSTER and TLD perform comparably to best short-term trackers (SiamFC and MDNet), while the CMT performs poorly due to a poor visual model.

The other two challenging attributes are fast motion and deformable object. Fast object motion is related to long-term re-detection, in both cases a large search range is beneficial (FCLT, SiamFC, MDNet). Deformable objects require quickly adaptable visual models, which is often in contradiction with the conservative updates that are required in long-term tracking.

The similar objects attribute shows the capability of handling multiple objects in the image. The performance here is similar to the performance on the short-term attributes since most of the trackers do not perform target re-detection on the whole image. The trackers which perform full-image re-detection have mechanism to prevent false detection of the similar targets, e.g., motion model in FCLT or they are not very successful in re-detecting due to the weak visual model like MUSTER and CMT.

\subsection{Tracking speed analysis}  \label{sec:speed-analysis}

\begin{figure}[h]
\begin{center}
	\includegraphics[width=0.9\linewidth]{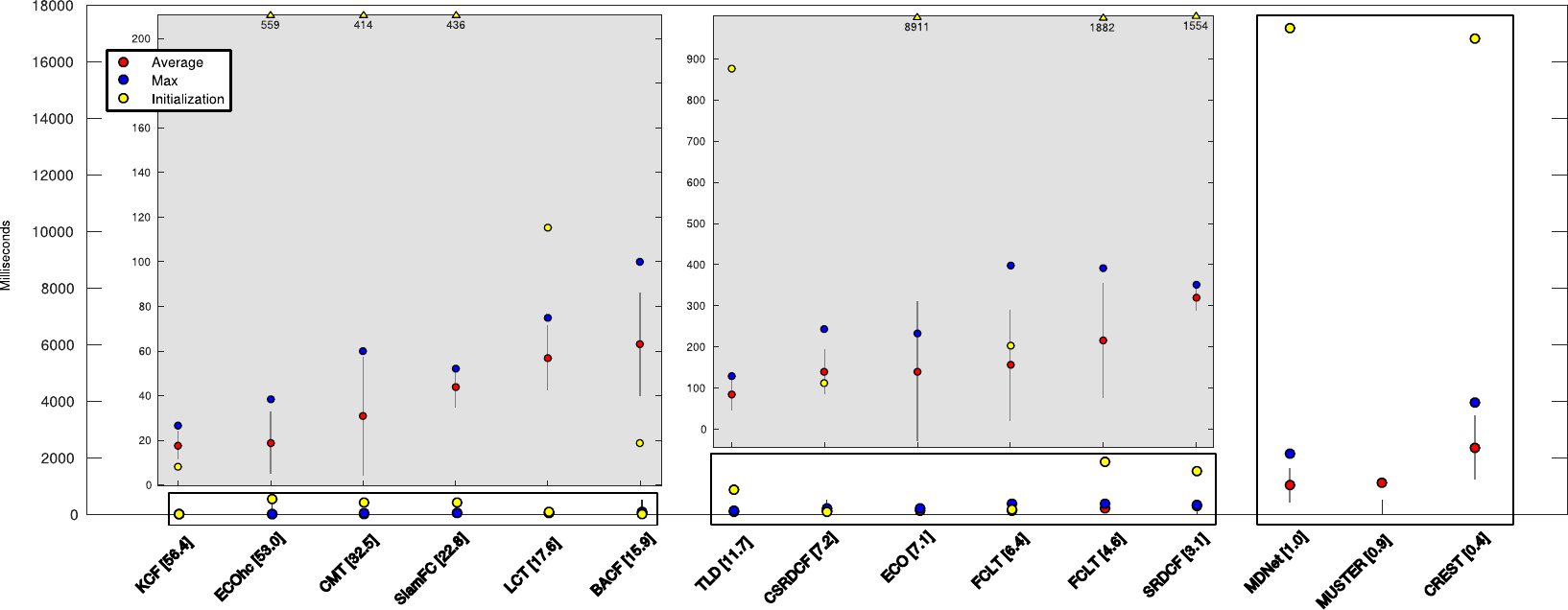}
\end{center}
   \caption{Speed performance of evaluated trackers. Trackers are ordered into three groups: fast (left), moderately fast (middle) and slow (right). All numbers are in milliseconds and an average speed frames-per-second is shown next to the name of each tracker.} 
\label{fig:time_analysis} 
\end{figure}

Tracking speed is a decisive factor in many applications. We provide a detailed
analysis by three measures: (i) initialization time, (ii) maximum per-frame time and (iii) average per-frame time. The initialization time is computed as the initial frame processing time averaged over all sequences. The maximal time is computed as the median of the slowest 10\% of the frames averaged over all sequences. The average time is averaged over all frames of the dataset. All measurements are in milliseconds per frame (MPF). The tracking speed is given in Figure~\ref{fig:time_analysis} with trackers categorized into three groups according to the average speed: fast ($>15$fps), moderately fast (1fps-15fps) and slow ($<1$fps).

The fastest tracker is the KCF due to efficient model learning and localization by fast Fourier transform. The slowest methods are CNN-based MDNet and CREST due to the time-consuming model adaptation and MUSTER due to slow keypoint extraction in detection phase. Several trackers exhibit a very high initialization time (in order of several thousand MPF). The delay comes from loading CNNs (SiamFC, ECO, PTAV, MDNet, CREST) or pre-calculating visual models (ECOhc, CMT, TLD, SRDCF).

Ideally, the tracking speed is approximately constant over all frames, which is reflected in small difference between the maximum per-frame and average time. This difference is largest for the following trackers: ECOhc and ECO (due to a time-consuming update every five frames), FCLT (due to re-detection on the entire image, which is slow for large images), PTAV (due to the slow CNN-based detector) and MDNet (due to the slow update during reliable tracking period).

\section{Conclusions and discussion}  \label{sec:conclusion}

The paper introduced a new long-term single-object tracking benchmark. We proposed a short-term/long-term taxonomy of visual trackers that predicts performance on sequences with long-term properties. New performance evaluation measures, sensitive to long-term aspects of tracking, were proposed as well. These measures offer significant  insights into long-term tracker performance and reduce to a standard short-term performance measures in a short-term tracking scenario, linking the two tracking problems.

We constructed a new dataset, the LTB35,  which focuses on target disappearances and emphasizes  long-term tracking attributes. Six long-term and nine short-term SOTA trackers were analyzed using the proposed methodology and the dataset. The overall evaluation, presented in Section~\ref{sec:overall_evaluation}, shows that the dataset is challenging, the best tracker achieves average F-score of 0.48, leaving room for improvement. Results show that, apart from efficient target re-detection mechanisms, careful updating of the visual model is crucial for dealing with long-term sequences. This is supported by the fact that several short-term trackers with conservative model updates perform well.

Results in Section~\ref{sec:sequence_evaluation} show that long-term tracking performance is not directly correlated with the sequence length, but rather with the number of target disappearances. 
This is further highlighted in the per-attribute analysis (Section~\ref{sec:attribute_evaluation}) and supports our approach to the  LTB35 dataset construction. Full occlusions and out-of-view disappearances are among the most challenging attributes. The other are fast motion (related to the search range of the tracker) and deformable targets which requires highly adaptive and robust visual model.

Tracking speed analysis experiments show that reporting solely average speed may be misleading and insufficient for applications that require short response times. In Section~\ref{sec:speed-analysis} we show that many trackers, especially long-term, perform very expensive re-detection or learning operations at regular or even unpredictable time instances. Furthermore, initialization times for several trackers are order of magnitude larger than the standard tracking iteration. We conclude that additional information, like the maximum response time and initialization times should be reported as part of standard analysis.

\clearpage

\bibliographystyle{splncs}
\bibliography{bib}

\begin{thebibliography}{10}

\bibitem{otb_cvpr2013}
Wu, Y., Lim, J., Yang, M.H.:
\newblock Online object tracking: A benchmark.
\newblock In: Comp. Vis. Patt. Recognition. (2013)  2411-- 2418

\bibitem{alov_pami2014}
Smeulders, A., Chu, D., Cucchiara, R., Calderara, S., Dehghan, A., Shah, M.:
\newblock Visual tracking: An experimental survey.
\newblock IEEE Trans. Pattern Anal. Mach. Intell. \textbf{36}(7) (July 2014)
  1442--1468

\bibitem{templecolor_tip2015}
Liang, P., Blasch, E., Ling, H.:
\newblock Encoding color information for visual tracking: Algorithms and
  benchmark.
\newblock IEEE Trans. Image Proc. \textbf{24}(12) (Dec 2015)  5630--5644

\bibitem{kristan_vot_tpami2016}
Kristan, M., Matas, J., Leonardis, A., Vojir, T., Pflugfelder, R., Fernandez,
  G., Nebehay, G., Porikli, F., Cehovin, L.:
\newblock A novel performance evaluation methodology for single-target
  trackers.
\newblock IEEE Trans. Pattern Anal. Mach. Intell. (2016)

\bibitem{MOTChallenge2015}
Leal-Taix\'{e}, L., Milan, A., Reid, I., Roth, S., Schindler, K.:
\newblock {MOTC}hallenge 2015: {T}owards a benchmark for multi-target tracking.
\newblock arXiv:1504.01942 [cs] (April 2015) arXiv: 1504.01942.

\bibitem{kristan_vot2017}
Kristan, M., Leonardis, A., Matas, J., Felsberg, M., Pflugfelder, R.,
  Cehovin~Zajc, L., Vojir, T., Hager, G., Lukezic, A., Eldesokey, A.,
  Fernandez, G.:
\newblock The visual object tracking vot2017 challenge results.
\newblock In: The IEEE International Conference on Computer Vision (ICCV).
  (2017)

\bibitem{otb_pami2015}
Wu, Y., Lim, J., Yang, M.H.:
\newblock Object tracking benchmark.
\newblock IEEE Trans. Pattern Anal. Mach. Intell. \textbf{37}(9) (Sept 2015)
  1834--1848

\bibitem{uav_benchmark_eccv2016}
Mueller, M., Smith, N., Ghanem, B.:
\newblock A benchmark and simulator for uav tracking.
\newblock In: Proc. European Conf. Computer Vision. (2016)  445--461

\bibitem{moudgil2017long}
Moudgil, A., Gandhi, V.:
\newblock Long-term visual object tracking benchmark.
\newblock arXiv preprint arXiv:1712.01358 (2017)

\bibitem{tao2017tracking}
Tao, R., Gavves, E., Smeulders, A.W.:
\newblock Tracking for half an hour.
\newblock arXiv preprint arXiv:1711.10217 (2017)

\bibitem{kristan_vot2013}
Kristan, M., Pflugfelder, R., Leonardis, A., Matas, J., Porikli, F., Čehovin,
  L., Nebehay, G., Fernandez, G., Vojir, T.e.a.:
\newblock The visual object tracking vot2013 challenge results.
\newblock In: Vis. Obj. Track. Challenge VOT2013, In conjunction with ICCV2013.
  (Dec 2013)  98--111

\bibitem{Galoogahi_2017_ICCV}
Kiani~Galoogahi, H., Fagg, A., Huang, C., Ramanan, D., Lucey, S.:
\newblock Need for speed: A benchmark for higher frame rate object tracking.
\newblock In: Int. Conf. Computer Vision. (2017)

\bibitem{cehovin_iccv2017}
Cehovin~Zajc, L., Lukezic, A., Leonardis, A., Kristan, M.:
\newblock Beyond standard benchmarks: Parameterizing performance evaluation in
  visual object tracking.
\newblock In: Int. Conf. Computer Vision. (2017)

\bibitem{cehovin_tip2016}
{\v{C}}ehovin, L., Leonardis, A., Kristan, M.:
\newblock Visual object tracking performance measures revisited.
\newblock IEEE Trans. Image Proc. \textbf{25}(3) (2016)  1261--1274

\bibitem{henriques2015tracking}
Henriques, J.F., Caseiro, R., Martins, P., Batista, J.:
\newblock High-speed tracking with kernelized correlation filters.
\newblock IEEE Trans. Pattern Anal. Mach. Intell. \textbf{37}(3) (2015)
  583--596

\bibitem{srdcf_iccv2015}
Danelljan, M., Hager, G., Shahbaz~Khan, F., Felsberg, M.:
\newblock Learning spatially regularized correlation filters for visual
  tracking.
\newblock In: Int. Conf. Computer Vision. (2015)  4310--4318

\bibitem{Lukezic_CVPR_2017}
Lukežič, A., Vojíř, T., Čehovin Zajc, L., Matas, J., Kristan, M.:
\newblock Discriminative correlation filter with channel and spatial
  reliability.
\newblock In: Comp. Vis. Patt. Recognition. (2017)  6309--6318

\bibitem{mdnet_cvpr2016}
Nam, H., Han, B.:
\newblock Learning multi-domain convolutional neural networks for visual
  tracking.
\newblock In: Comp. Vis. Patt. Recognition. (June 2016)  4293--4302

\bibitem{danelljan_iccv2015_convolutional}
Danelljan, M., Häger, G., Khan, F.S., Felsberg, M.:
\newblock Convolutional features for correlation filter based visual tracking.
\newblock In: IEEE International Conference on Computer Vision Workshop
  (ICCVW). (Dec 2015)  621--629

\bibitem{kalal_pami}
Kalal, Z., Mikolajczyk, K., Matas, J.:
\newblock Tracking-learning-detection.
\newblock IEEE Trans. Pattern Anal. Mach. Intell. \textbf{34}(7) (July 2012)
  1409--1422

\bibitem{Pernici2013}
Pernici, F., Del~Bimbo, A.:
\newblock Object tracking by oversampling local features.
\newblock IEEE Trans. Pattern Anal. Mach. Intell. \textbf{36}(12) (2013)
  2538--2551

\bibitem{CMT_CVPR2015}
Nebehay, G., Pflugfelder, R.:
\newblock Clustering of static-adaptive correspondences for deformable object
  tracking.
\newblock In: Comp. Vis. Patt. Recognition. (2015)  2784--2791

\bibitem{Maresca2013}
Maresca, M.E., Petrosino, A.:
\newblock Matrioska: A multi-level approach to fast tracking by learning.
\newblock In: Proc. Int. Conf. Image Analysis and Processing. (2013)  419--428

\bibitem{muster_cvpr2015}
Hong, Z., Chen, Z., Wang, C., Mei, X., Prokhorov, D., Tao, D.:
\newblock Multi-store tracker (muster): A cognitive psychology inspired
  approach to object tracking.
\newblock In: Comp. Vis. Patt. Recognition. (June 2015)  749--758

\bibitem{LCT_CVPR2015}
Ma, C., Yang, X., Zhang, C., Yang, M.H.:
\newblock Long-term correlation tracking.
\newblock In: Comp. Vis. Patt. Recognition. (2015)  5388--5396

\bibitem{ptav_iccv2017}
Fan, H., Ling, H.:
\newblock Parallel tracking and verifying: A framework for real-time and high
  accuracy visual tracking.
\newblock In: Int. Conf. Computer Vision. (2017)  5486--5494

\bibitem{fclt_arxiv}
Lukezic, A., Zajc, L.C., Voj{\'{\i}}r, T., Matas, J., Kristan, M.:
\newblock {FCLT} - {A} fully-correlational long-term tracker.
\newblock CoRR \textbf{abs/1711.09594} (2017)

\bibitem{everingham2010pascal}
Everingham, M., Van~Gool, L., Williams, C.K., Winn, J., Zisserman, A.:
\newblock The {PASCAL} visual object classes ({VOC}) challenge.
\newblock International journal of computer vision \textbf{88}(2) (2010)
  303--338

\bibitem{DanelljanCVPR2017}
Danelljan, M., Bhat, G., Shahbaz~Khan, F., Felsberg, M.:
\newblock Eco: Efficient convolution operators for tracking.
\newblock In: Comp. Vis. Patt. Recognition. (2017)  6638--6646

\bibitem{BACF_ICCV2017}
Kiani~Galoogahi, H., Fagg, A., Lucey, S.:
\newblock Learning background-aware correlation filters for visual tracking.
\newblock In: Int. Conf. Computer Vision. Number 1135--1143 (2017)

\bibitem{siamfc_eccv16}
Bertinetto, L., Valmadre, J., Henriques, J.F., Vedaldi, A., Torr, P.H.:
\newblock Fully-convolutional siamese networks for object tracking.
\newblock (2016)

\bibitem{crest_ICCV17}
Song, Y., Ma, C., Gong, L., Zhang, J., Lau, R.W.H., Yang, M.H.:
\newblock Crest: Convolutional residual learning for visual tracking.
\newblock In: The IEEE International Conference on Computer Vision (ICCV).
  (2017)

\bibitem{sint_cvpr16}
Tao, R., Gavves, E., Smeulders, A.W.M.:
\newblock Siamese instance search for tracking.
\newblock In: Proceedings of the IEEE Conference on Computer Vision and Pattern
  Recognition. (2016)

\bibitem{kristan_vot2016}
Kristan, M., Leonardis, A., Matas, J., Felsberg, M., Pflugfelder, R.,
  \v{C}ehovin, L., Vojir, T., H\"{a}ger, G., Lukežič, A., et~al. Fernandez,
  G.:
\newblock The visual object tracking vot2016 challenge results.
\newblock In: Proc. European Conf. Computer Vision. (2016)

\end{thebibliography}
\end{document}